\documentclass[journal]{IEEEtran}
\usepackage{graphicx}
\usepackage{subfigure}
\usepackage{array}
\usepackage{amsmath,epsfig}
\usepackage{booktabs}
\usepackage[ruled]{algorithm2e}
\usepackage{algorithmic}
\usepackage{algorithm2e}
\usepackage{multirow}
\usepackage{diagbox}
\usepackage[table]{xcolor}
\usepackage{amssymb}
\usepackage[driverfallback=dvipdfm,
CJKbookmarks=true, bookmarksnumbered=true, bookmarksopen=true,
colorlinks=true,
pdfborder=001,
citecolor=blue, linkcolor=blue, anchorcolor=red, urlcolor=black
]{hyperref}
\begin{document}\sloppy
\title{Attention Consistency Refined Masked Frequency Forgery Representation  for Generalizing Face Forgery Detection}

\author{Decheng~Liu,~Tao~Chen,~Chunlei~Peng,~\IEEEmembership{Member, IEEE},~Nannan~Wang,~\IEEEmembership{Member, IEEE},~Ruimin~Hu,~Xinbo~Gao,~\IEEEmembership{Senior Member, IEEE}
}


\markboth{Journal of \LaTeX\ Class Files,~Vol.~16, No.~8, July~2023}%
{Shell \MakeLowercase{\textit{et al.}}: A Sample Article Using IEEEtran.cls for IEEE Journals}


\IEEEcompsoctitleabstractindextext{%
\begin{abstract}

Due to the successful development of deep image generation technology, visual data forgery detection would play a more important role in social and economic security. Existing forgery detection methods suffer from unsatisfactory generalization ability to determine the authenticity in the unseen domain. 
In this paper, we propose a novel Attention Consistency Refined masked frequency forgery representation model toward generalizing face forgery detection algorithm (ACMF). 
Most forgery technologies always bring in high-frequency aware cues, which make it easy to distinguish source authenticity but difficult to generalize to unseen artifact types.
The masked frequency forgery representation module is designed to explore robust forgery cues by randomly discarding high-frequency information. 
In addition, we find that the forgery attention map inconsistency through the detection network could affect the generalizability. 
Thus, the forgery attention consistency is introduced to force detectors to focus on similar attention regions for better generalization ability. 
Experiment results on several public face forgery datasets (FaceForensic++, DFD, Celeb-DF, and WDF datasets) demonstrate the superior performance of the proposed method compared with the state-of-the-art methods. 
\emph{The source code and models are publicly available at \href{https://github.com/chenboluo/ACMF}{https://github.com/chenboluo/ACMF}.}

\end{abstract}

\begin{keywords}
Forgery detection, generalization, frequency domain, network visualization.
\end{keywords}}

\maketitle
\IEEEdisplaynotcompsoctitleabstractindextext
\IEEEpeerreviewmaketitle

\section{Introduction}
With the rapid development of deep generative networks, more and more photo-realistic images or videos are produced and spread on social media~\cite{groshev2022ghost,jia2022video}.
Especially, advanced face artifact technology brings a high risk of malicious abuse in existing biometric systems to threaten social and economic security.
Face forgery detection~\cite{zhang2017automated} refers to precisely distinguishing between forged and real faces drawing more and more attention to computer vision and security fields.
The key issue of face forgery detection is to explore robust and discriminative artifact clues in forged images, which guarantees that it can be applied in real complex scenarios.

\begin{figure}[t]
  \centering
  \includegraphics[width=\linewidth]{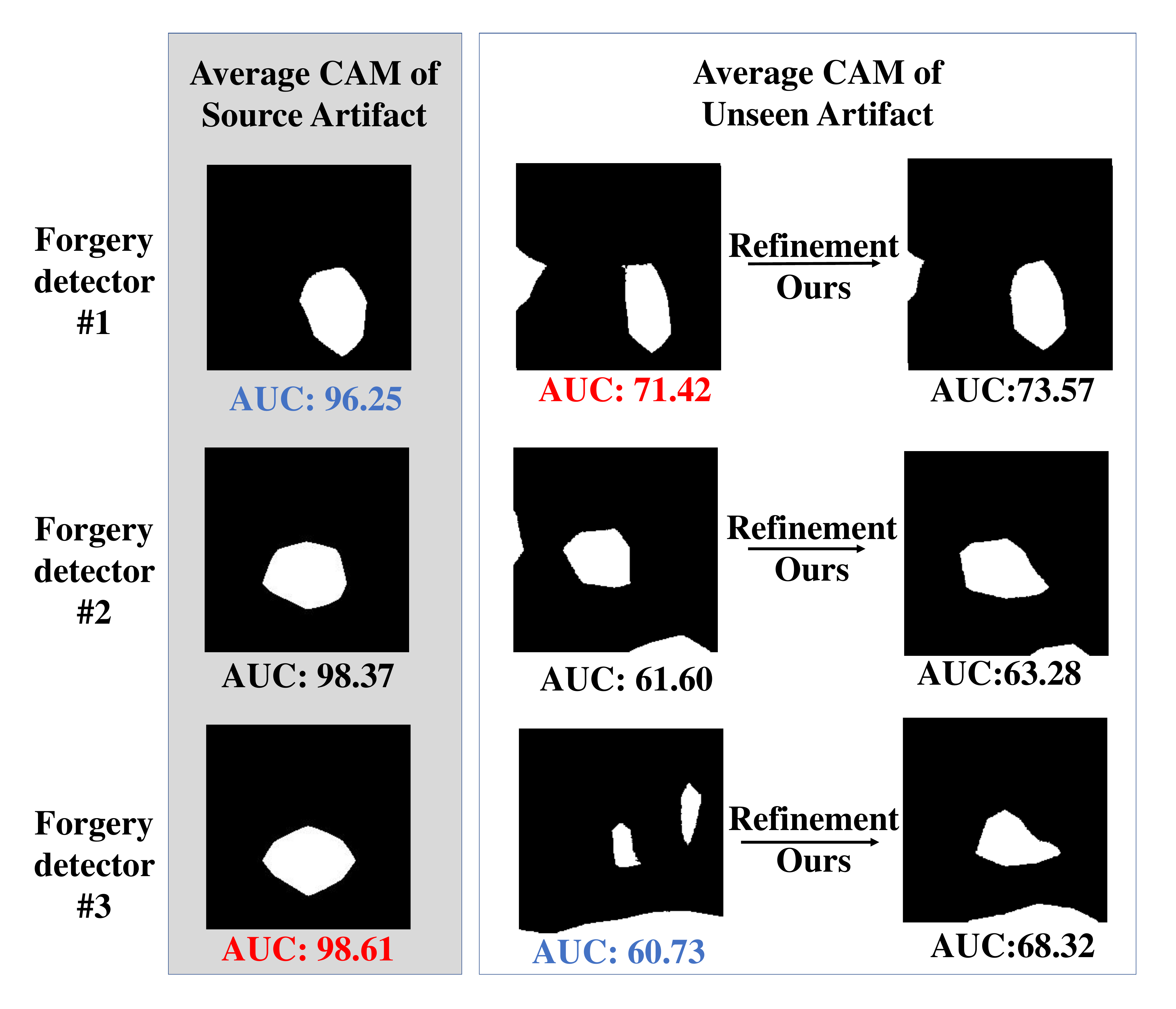}

  \caption{
  A visual illustration of average class activation maps of different forgery detectors trained on the FF++ dataset.
  Specifically, we calculate the CAM of dataset samples and then average it.
  It can be found the forgery detector models with good performance in intra-dataset evaluation are prone to perform poorly in cross-dataset evaluation.
  And the forgery attention map inconsistency can affect the generalizability.
  Inspired by the phenomenon, we design the cross forgery attention consistency to constraint similar attention maps and improve performance.
  The red color denotes the highest accuracy and blue color denotes the lowest accuracy.
  }\label{fig:first}

\end{figure}

Early related works focus on designed hand-crafted features~\cite{zhang2017automated} to analyze the difference between real and fake images.
Recently, most researchers consider the forgery detection task as the image binary classification task, and deep learning-based algorithms~\cite{he2019detection,zhang2019detecting} are proposed to distinguish forged faces.
Limited by the interpretability of deep networks and collecting data scale, most existing methods suffer from the serious overfitting problem.
They could achieve almost 99\% accuracy in intra-dataset evaluation scenarios~\cite{ni2022core} but fail to distinguish fake data generated by unseen methods~\cite{chollet2017Xception}. 
Thus, the generalizability becomes a key criterion for judging the forgery detection model. 
Previous researches show it is an effective way to mine face forgery clues in the frequency domain~\cite{dong2023contrastive,wang2023frequency}.
It is because the high-frequency component contains complex contours, textures, and edges in images and even establishes a strong dependence on the forgery clues.
However, existing methods pay much attention to leveraging high-frequency information to boost accuracy performance, ignoring the generalizability to mimic real-world scenarios.

The network visualization technologies~\cite{selvaraju2017grad,wang2021representative} can provide an interpretable perspective for humans to explore the underly rules to help recognize artifact faces through meticulously designed experiments.
\cite{selvaraju2017grad} gives a visualization tool to analyze the attention region in the designed representation model.
Figure 1 shows that forgery detector models with high performance in intra-dataset evaluation are prone to perform
poorly in cross-dataset evaluation. We assume that the forgery attention map inconsistency can affect the generalizability. 
Additionally, the forgery detection model with more inconsistency of the forgery attention map between testing data and training data would result in poorer cross-dataset performance.
This phenomenon (as illustrated in Figure 1) naturally inspires us to improve generalizability by designing attention consistency regularization to refine the former forgery detection network focusing on similar spatial attention regions.

In this paper, we propose a novel attention consistency refined masked frequency forgery representation model algorithm toward generalizing face forgery detection.
To alleviate the overfitting problem, the designed face forgery representation aims to discover more difficult forgery clues in the frequency domain, which are partly erased to boost the generalization ability to unseen forged types.
Furthermore, the attention consistency constraint is leveraged to refine the consistency of the network attention region in an adaptive strategy.
Specifically, we first decouple the high-frequency component and low-frequency component for inputting forgery face images through Fourier transform processing.
Then, we design the masked frequency forgery representation model module to partly erase high-frequency information for generalizability.
Finally, the proposed cross-attention consistency regularization is leveraged to guide the detector to refine and align its attention consistency, which could further boost generalization performance.

The main contributions of our paper can be summarized as follows:

\begin{enumerate}
    \item The masked frequency forgery representation module is designed to partly discard sensitive high-frequency information, which can help explore more robust forgery clues in the frequency domain for unseen artifacts to mimic real-world scenarios.
    
    \item We find that the source forgery attention map inconstancy can affect the generalizability of face forgery detection models. To this end, we design the forgery attention consistency regularization to guide the forgery detector to refine and align its attention region.

    \item Experimental results on several public face forgery datasets illustrate the superior performance of the proposed ACMF compared with the state-of-the-art face forgery detection methods. 
    \emph{The code is publicly available at \href{https://github.com/chenboluo/ACMF}{https://github.com/chenboluo/ACMF}}.
\end{enumerate}

We organized the rest of this paper as follows. Section 2 shows some representative face forgery detection algorithms. 
In Section 3, we present the novel attention consistency refined masked frequency forgery representation model toward generalizing face forgery detection algorithm. 
Section 4 shows the experimental results and analysis. The conclusion is drawn in Section 5. 

\section{Related work}

\subsection{Frequency Forgery Detection}
The frequency forgery detection is defined by forgery detection in frequency domain.
The early frequency face forgery detection task focuses on developing models that can detect mainly known forgeries. These models can be classified into two categories: those that use spatial cues and those that use temporal cues for detection.

\textbf{
Spatial cues based frequency face forgery detection.} Drawing on traditional face recognition methods, Zhang \emph{et al.}~\cite{zhang2017automated} pioneered a face forgery detection technique, which uses multiple classifiers for discriminating forged images after feature extraction. Based on this, subsequent researchers identified some visual detail mismatches in face forgery images and used them for detection. For example, He \emph{et al.}~\cite{he2019detection} combined a random forest classifier to identify forged images based on image residual information in different color spaces. The literature~\cite{zhang2019detecting} reveals that the forged images generated by GAN models have a special distortion feature and distinguish them in the frequency domain spectrogram. Luo \emph{et al.}~\cite{mo2018fake} preprocessed the images by high-pass filters to enhance the detection. Dong \emph{et al.}~\cite{xuan2019generalization} proposed the use of image preprocessing such as Gaussian blur and noise to eliminate the difference of high-frequency distortion between the forged and real images, so as to better learn the forgery features and improve the detection performance.

\textbf{Temporal cues based frequency face forgery detection.}
A common flaw in forged videos is the discontinuity between video frames, which can be used as a basis for detecting forged videos. Researchers at Purdue University~\cite{guera2018deepfake} first used this cue to detect forged videos. They used a convolutional neural network to extract the features of each frame, then a recurrent neural network to capture the temporal features, and finally a full convolutional neural network to classify whether the video is a forgery. a similar scheme combining convolutional and recurrent neural networks was used by Sabir \emph{et al.}~\cite{sabir2019recurrent} to detect forged videos. Amerini \emph{et al.}~\cite{amerini2019deepfake}, on the other hand, proposed an optical flow method to compute the inter-frame differences and feeding them into a convolutional neural network for fake video detection. Masi \emph{et al.}~\cite{masi2020two} amplified artifacts in video streams to detect fake faces and suppress advanced face content. Wu \emph{et al.}~\cite{9053969} extracted hidden analysis features on convolutional filters for detect hidden tampered artifacts.

Most of these frequency-based forgery detection methods used all frequency information as a significant forgery cue. So these methods perform well on test videos or images which contain the same forgery type as the training data. However, due to the obvious frequency cues of the existing forgery dataset, the previously discovered forgery laws often fail when detecting different kinds of forged faces. Therefore, we propose to explore robust high-frequency forgery cues to obtain good generalization ability.


\begin{figure*}[h]
  \centering
  \includegraphics[width=0.8\linewidth]{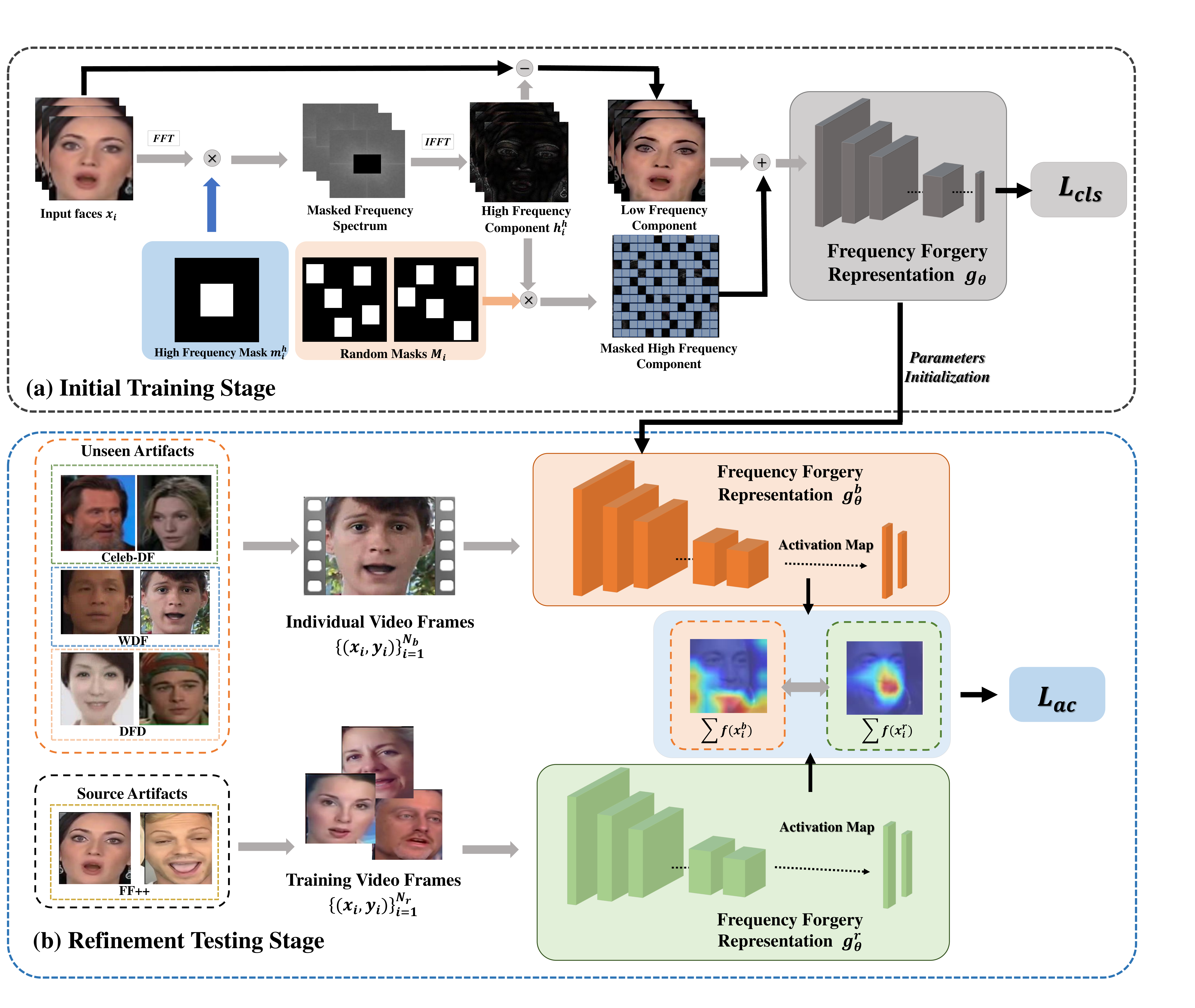}
  \caption{
  Overview of the proposed attention consistency refined masked frequency forgery representation for generalizing face forgery detection (ACMF).
  The ACMF contains two stages: the MFR learning training stage and the FAC refinement testing stage. We adjust the weight of the model on a per-video basis to improve generalizability.
 } \label{fig:overview}
\end{figure*}

\subsection{Generalized Forgery Detection}

Generalized forgery detection is defined by distinguishing unseen forgery types in the cross-dataset evaluation.
Existing most face forgery methods often fail to detect unseen artifacts. 
To address this problem, many researchers have leveraged deep learning model and  data enhancements to increase the generalization performance.

\textbf{Deep learning model based generalized forgery detection.}
He \textit{et al.}~\cite{he2021beyond} developed a strategy that integrates super-resolution, denoising and shadow reorganization techniques to extract robust features and residual visual cues for detecting forged images. 
Miao \textit{et al.}~\cite{miao2021towards}, inspired by the bag-of-words model in natural language processing, proposed to decompose images into local feature vectors for forgery detection, which enhances the network’s efficiency in detecting local anomalies.
Xu \textit{et al.}~\cite{xu2022supervised} leveraged the diversity of different network architectures and fused their outputs. They applied contrast loss to identify unknown types of depth forgery and improved the consistency of their results. Lin \textit{et al.}~\cite{lin2022improved} used convolution layers at different levels to capture high-level semantic features of face images and performed feature fusion, refinement and reorganization to achieve better generalization performance than the baseline method.  Wang \textit{et al.}~\cite{wang2022adt} proposed a transformation-based framework for modeling and analyzing global and local information of facial images. The method relies less on local texture features in the training data, which increases its generality.

\textbf{Data enhancement based generalized forgery detection.}
Some researchers have explored using noise to mask facial recognition features and extract forgery cues that are hard to tamper with. Wang \textit{et al.}~\cite{wang2022deepfake} applied pixel-level Gaussian blurring to diminish the effect of high-frequency forgery features and incorporated adversarial training into the training of a deep forgery detection model, which significantly enhanced the model’s generalization ability and robustness. Guo \textit{et al.}~\cite{guo2021fake} devised an adaptive forgery trace extraction network that can serve as a preprocessing step to eliminate image content and emphasize manipulation traces. Based on this, Wang \textit{et al.}~\cite{nadimpalli2022improving} proposed to remove structured information region features. Guo \textit{et al.}~\cite{guo2023data} segmented the candidate face images into strongly and weakly correlated regions and designed a MSF module that boosts the detector’s robustness to these regions by dynamically erasing some locations.

However, existing deep learning model based methods are poor scalability and lack interpretability.
Data enhanced based methods can't effectively leverage the specific characteristics of different individual forged videos.
Considering the phenomenon as illustrated in Figure 1, we propose cross attention consistency regularization to refine detector to boost the generalizaiton ability.

\section{Methodology}

\subsection{Preliminaries}
In this paper, we aim to design a novel Attention Consistency refined masked frequency forgery representation model method (ACMF), which could mine robust forgery clues in the frequency domain and be restricted to cross-dataset attention consistency. 
The basic intuitions behind our proposed generalized forgery detection method are as follows: 
\textit{
(1) existing face manipulation technologies often bring in high-frequency cues which are easily distinguished but difficult to generalize to unseen artifact types; 
(2) the forgery attention map inconsistency through the detection network would affect the generalizability (as illustrated in Figure 1).
}
Toward this end, we propose the masked frequency forgery representation model (MFR) to discard high-frequency artifacts and learn more generalizable forgery cues.
Inspired by the interesting phenomenon, we further design the cross-forgery attention consistency (FAC), which can be regarded as the refining strategy to boost generalizability.
As shown in Figure 2, the training procedure can be divided into two stages, and more details will be expressed in the following.

Given a forgery detection dataset denoted as $\left\{ {({x_i},{y_i})} \right\}_{i = 1}^N$, where ${x_i} \in {R^{H \times W \times 3}}$ denotes the input RGB face image, $H$, and $W$ separately represents the height and width of the image, ${y_i} \in \{ 0,1\}$ means the forgery detection could be regarded as the binary classification task,
and $N$ is the total number of training frames or images. 
The goal of the proposed masked frequency forgery representation is to train robust discriminative features ${g_\theta}({x_i})$, where ${g_\theta}$ means the forgery representation network mapping function.

\subsection{Masked Frequency Forgery Representation}

Considering that conspicuous high-frequency artifacts are prone to affect the generalizability of the forgery detection model, we propose the masked frequency forgery representation by randomly discarding redundant frequency information as illustrated in Figure 2.
Here we choose the Xception~\cite{chollet2017Xception} model as the backbone network to explore discriminative forgery cues, and masked frequency information strategy could boost the generalization ability.
Firstly, we use the two-dimension discrete Fourier transform algorithm to transform $x_i$ from the RGB image domain to the frequency domain as follows:
\begin{equation}
{z_i} = {F}\left( {{x_i}} \right),
\end{equation}
where $ {F}( \cdot )$ denotes the discrete fourier transform, and ${{F}^{ - 1}}( \cdot )$ denotes the reverse discrete fourier transform.
It can be found that the low frequency information is placed in the  center and the high frequency information is placed in the edge after specific processing.
Then, we decouple the high frequency component and low frequency component by masking frequency information.
Assuming the conner of the Fourier spectrum as the original point, we define the distance in the frequency domain to set boundary between low frequency component and high frequency component as follow:
\begin{equation}
Dis(u,v) = max(\left| {u - \frac{H}{2}} \right|,\left| {v - \frac{W}{2}} \right|),
\end{equation}
where $ (u,v)$ is the position of the spectrum. 
Following, we set the threshold $d$ to make boundary to divide different frequency components.
When distances $ Dis(u,v)$ are higher than threshold are denoted as high frequency components, and distances $ Dis(u,v)$ are lower than threshold are denoted as low frequency components.
Thus, we can calculate the high frequency component $h_i^h$ by multiplying a rectangular mask $m_i^h$ as follows:
\begin{equation}
h_{_i}^h = {F^{ - 1}}({z_i} \times m_i^h),
\end{equation}
\begin{equation}
h_{_i}^l = x_i - {F^{ - 1}}({z_i} \times m_i^h),
\end{equation}
where the value of the rectangular mask $m_i^h$ is set to 1 when ${{\rm{Dis(u,v)  >  d}}}$, the value of the rectangular mask $m_i^h$ is set to 0 when $ {Dis(u,v) \le d}$.
Here the parameter $d$ is important to divide high frequency component and more analysis is described in the following section.
Inspired by \cite{MAE}, we design a random mask to discard high frequency information following.
Thus, the processed face images can be calculated with randomly masked high frequency component and source low frequency component.
\begin{equation}
{\widehat x_i} = h_{_i}^h \times {M_i} + h_{_i}^l,
\end{equation}
where the radio $r$ is an important parameter of the random mask $M_i$ and related analysis is expressed in experiment section. 

Finally, we fed these processed images $ {\widehat x_i}$ into the classifier network to train parameters of the designed MFR model. 
For convenience of analysis, we utilize the cross entropy between the label $y_i$ and the output probability as follows:
\begin{equation}
{L_{{\rm{cls}}}} =  - \sum\limits_{i = 1}^N {\sum\limits_{k = 0}^1 {{y_{i,k}}\log ({H_\phi }({\widehat x_i},{g_\theta }))} } ,
\end{equation}
where $ {{H_\phi }}$ means the classifier head implemented with a softmax function, and $ {{H_\phi }({\widehat x_i},{g_\theta })}$ is the prediction vector.


\begin{algorithm}[!t]
	\caption{ The proposed ACMF algorithm}\label{algorithm}
	\KwIn{Training data $X=\left\{x_1,...,x_n\right\},Label=\left\{y_1,...,y_n\right\}$ and testing data $Z=\left\{z_1,...,z_m\right\}$. Total number of training rounds in the MFR training stage is set as $R$.
 And total number of training  rounds in the FAC Refinement stage is set as $T$. Random mask ratio is $r$ in MFR module.
 The forgery representation network with trained parameters is denoted as $g_\theta$}.
    \KwOut{Probability of forgery for each test video $y_{pred}=\left\{y_{pred}^{1},...,y_{pred}^{m}\right\}$}
    
	\begin{algorithmic}[1]
        \STATE Initialize forgery detector $g_\theta$
        \WHILE{$ r < R $}
            \STATE Read mini-batch from training set $X$;
            \STATE Randomly discarding high frequency information to construct image ${\widehat x_i}$ for each sample $x_i$ via Eq. 5;
            \STATE Calculate $L_{cls}$ via Eq. 6;
            \STATE  $Update  \quad  g_\theta^0$\\

        \STATE $ r \gets r+1 $
        \ENDWHILE
        
        \FORALL{$Z_{i} \in Z$}
            \STATE $g_\theta^1 \gets g_\theta^0$    
            \STATE $t \gets 1$
            \WHILE{$ t < T $}
                \STATE Calulate forgery attention map for $x_i$;
                \STATE Calculate $L_{ac}$ via Eq. 7;
                \STATE  $Update  \quad  g_\theta^1$         
                \STATE  $t \gets t+1 $               
            \ENDWHILE
            \STATE $y_{pred}^{i} \gets g_\theta^1(Z_i)$
        \ENDFOR        
	\end{algorithmic} 
\end{algorithm}

\subsection{Forgery Attention Consistency}

Motivated from the novel phenomenon as illustrated in Figure 1, we aim to force the learned forgery detector network focus on similar attention regions to improve generalizability.
Thus, we further design the forgery attention consistency to refine mentioned MFR model for performance improvement.
Here we choose the Grad-CAM visualization algorithm~\cite{selvaraju2017grad} and refer to the work of Lsemi~\textit{et al.}~\cite{lee2021semi} to determine which spatial regions are sensitive to the classifier, where perturbation would bring the most significant impact on performance.
We denote the forgery attention map as $ {\rm{f}}({x_i}) \in {R^{H \times W}}$, which contains the same size of input image.
For each individual input video, the cross forgery attention consistency is formulated as:
\begin{equation}
{L_{ac}} = \;||\frac{1}{{{N_b}}}\sum\limits_{i = 1}^{{N_b}} {f({x_i}) - } \frac{1}{{{N_r}}}\sum\limits_{j = 1}^{{N_r}} {f({x_j})} ||_2^2,
\end{equation}
where the $N_b$ is the number of the input individual video frames, and $N_r$ is the number of the reference frames. Here we set the training data as reference frames to improve the generalizability. 
Detailed settings analysis is presented in the following section, and the algorithm of our method is shown in Algorithm 1.



\section{Experiment}
In this section, we first explore how different parameter settings affect the forgery detection performance. Then we conduct comprehensive experiments to demonstrate the superiority and effectiveness of our proposed method. Finally, we present rich visualizations of the results.

\subsection{Datasets}
To evaluate our method, we select several famous and challenging face forgery datasets, which include FaceForensics++ (FF++)~\cite{rossler2019faceforensics++}, DFD, Celeb-DF~\cite{li2020celeb}, and WDF~\cite{zi2020wilddeepfake}.

\begin{figure}
\begin{center}
\includegraphics[width=1\linewidth]{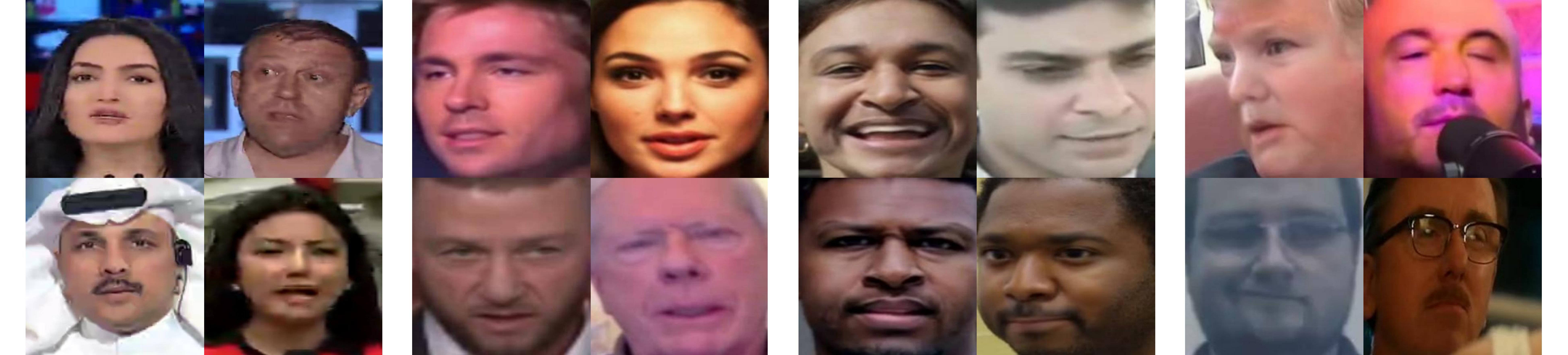} 
\begin{minipage}[b]{1\linewidth}
\centering
(a) \quad \quad \quad \quad \quad (b) \quad \quad  \quad \quad \quad (c) \quad \quad \quad \quad \quad (d)
\end{minipage}

\caption{The illustration of public representative forgery face datasets. (a) the FF++ dataset. (b) Celeb-DF dataset. (c) the DFD dataset. (d) the WDF dataset.}
\label{fig:dataset}
\end{center}  
\end{figure}

$\bullet$ FF++ dataset consists of 4,000 face forgery videos generated by four face forgery methods (DeepFake, Face2Face, FaceSwap, and NeuralTextures) and 1,000 original video sequences. With the same protocol~\cite{rossler2019faceforensics++}, we split FF++ datasets into three subsets, which include the training set consisting of 720 videos, the validation set consisting of 140 videos, and the test set consisting of 140 videos.

$\bullet$ DFD dataset records hundreds of real videos and creates thousands of deepfakes from these videos, Using publicly available deepfake generation methods. DFD dataset contains over 3,000 manipulated videos from 28 actors in various scenes. With the same protocol~\cite{rossler2019faceforensics++}, we select 140 videos for testing.

$\bullet$ Celeb-DF dataset is a widely used dataset, which contains 590 original videos from YouTube and 5,639 corresponding deepfake videos. Since the usage of the improved deepfake algorithm, the Celeb-DF dataset has no particularly obvious forgery clues compared to FF++. 
With the same protocol~\cite{li2020celeb}, we only use the test set of the Celeb-DF dataset for testing.

$\bullet$ WDF dataset contains 3,805 real videos and 3,509 fake videos, which are all collected from the Internet. So we cannot determine the forgery method used in the WDF dataset. Meanwhile, videos in WDF dataset are compressed. 
With the same protocol~\cite{zi2020wilddeepfake}, we only use the test set of the WDF dataset for testing.

\subsection{Implementation Details}
We extract frames from each video and process face alignment by retinaface~\cite{deng2020retinaface}. For the stability of the results, we select 50 frames in videos for the test.
 We resize all face images into $224\times 224$.
 Our proposed method is based on the Xception network~\cite{chollet2017Xception}. The Xception network consists of 14 modules, all of which have linear residual connections around them, except for the first and last modules.
 We use Adam optimizer to train the framework with betas of $0.9$ and $0.995$. The batch size of our experiment is $64$. We set the learning rate as $lr=0.001$. All models are tested and trained on two NVIDIA GeForce RTX 3090 24GB GPUs. To evaluate the proposed method, we report the area under the receiver operating characteristic curve (AUC).

\subsection{Parameter Analysis}
The parameters that appeared in this paper are set as follows. We conduct the parameter study on the FF++ dataset, Celeb-DF dataset, and WDF dataset to evaluate the effect of forgery detection performance. We select frames as the unit for the Parameter Analysis. To eliminate other confounding factors, we do not include the forgery attention consistency module during parameter Analysis.

\textbf{Influence of parameter $d$.}
We evaluate the effect of parameter $d$ in the FF++ dataset and WDF dataset.   In Table~\ref{table:d}, the effect of parameter $d$  from a set of $\left\{0, 6, 25, 50, 75, 95, 112\right\}$ is illustrated when other parameter values are fixed. $d$, which separates the high and low-frequency components of the image, which is important for forgery detection performance in the MFR module. It can be noticed that when the AUC of the inter-dataset test decreases as $d$ decreases. Meanwhile, with a decrease of $d$, the AUC of the cross-dataset test increases as a whole. We recommend maximizing the performance in the cross-dataset without compromising the performance in the inner-dataset significantly. Therefore, we select $d=5$ as our hyperparameter.

\begin{table}
  \caption{Parameter experiments of $d$. We set the value of $r$ to 50\%, and iterate over the value of $d$. The best AUC(\%) results for each group are shown in bold numbers.}\label{table:d}
\centering
  \label{tab:exp-d}
\begin{tabular}{|c|c|c|} 
  \hline
    \multirow{2}{*}{Value of $d$}   
    & Source & Unseen  \\ 
   \cline{2-3}
 & FF++  & WDF               \\ 
  \hline
112       & \textbf{96.3 }      & 62.72             \\ 
95        & 95.4     & 58.24             \\ 
75        & 95.97     & 63.93             \\ 
50        & 94.65   & 66.97             \\ 
25        & 94.98    & 67.37             \\ 
5         & 93.26    & \textbf{71.53}             \\ 
0         & 69.62    & 59.15             \\
\hline
\end{tabular}
\end{table}

\textbf{Influence of parameter mask ratio $r$.} 
We evaluate the effect of parameter mask ratio $r$ of the random mask $M_i$ in the FF++ dataset, Celeb-DF dataset, and WDF dataset.
In Table~\ref{table:mask}, the effect of parameter $r$ from a set of $0.0$ to $0.9$ is illustrated when other parameter values are fixed. $r$, which determines the number of high frequency components, which is important for forgery detection performance in the MFR module. 
It can be noticed that when the AUC of the inter-dataset test increases as $r$ decreases. Meanwhile, with the decrease of $r$, the AUC of the WDF dataset decreases, and the AUC of the Celeb-DF dataset firstly increases and then decreases. The celeb-DF dataset has several deepfake cues similar to forgery cues in the FF++ dataset. Therefore, reducing high-frequency components to a certain degree can reduce the model's dependence on other irrelevant or disturbing information, but too many high-frequency components losing will lead the model to lose some useful or auxiliary information. The WDF dataset loses a lot of high-frequency information in the process of video compression. Removing most of the high-frequency components can increase the model's sensitivity and utilization of the global inconsistency or discordance in the low-frequency components and improve the model's generalization ability for face forgery detection.

\begin{figure*}
\begin{center}

\includegraphics[width=1\linewidth]{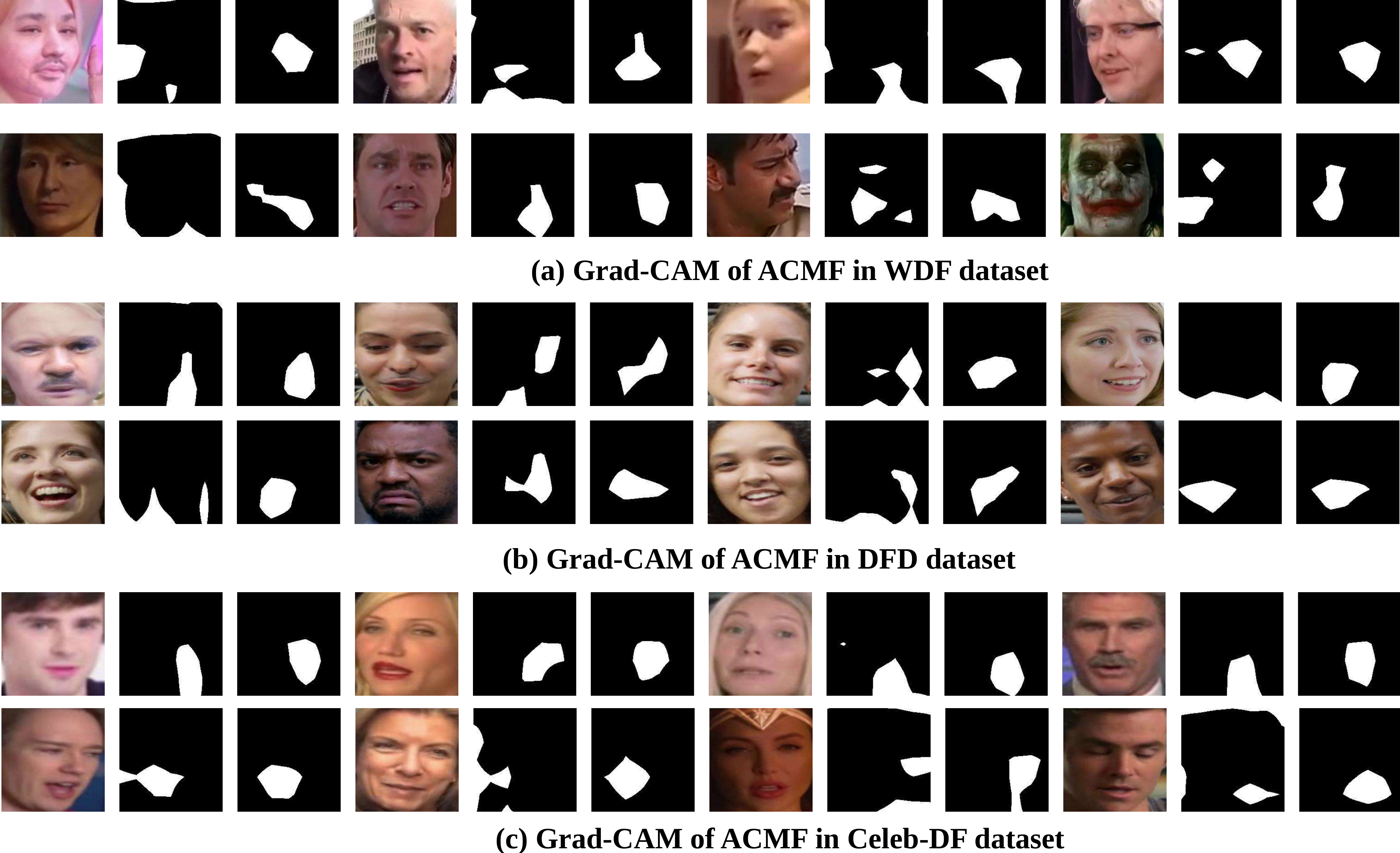}
\caption{
Grad Class activation maps (Grad-CAM) of ACMF in different datasets. Images are overlayed with the important attention regions highlighted with Grad-CAM~\cite{selvaraju2017grad}. We binarize the top 10\% of the forged pixels in the maps. (A) is the Grad-CAM of ACMF in WDF dataset. (B) is the Grad-CAM of ACMF in DFD dataset. (C) is the Grad-CAM of ACMF in Celeb-DF dataset.
It noted that the left subfigure is the input image, the middle subfigure shows the visualization of the Baseline, and the right subfigure shows the visualization of the proposed method ACMF.  }
\label{heatmap1}
\end{center}  
\end{figure*}

\begin{table}
\caption{Parameter experiments of mask ratio $r$. We set the value of $d$ to 5 and iterate over the value of $r$. The best AUC(\%) results for each group are shown in bold numbers.}\label{table:mask}
\centering
\begin{tabular}{|c|c|c|c|} 
\hline
    \multirow{2}{*}{Value of maskr ratio $r$}   &  Source & \multicolumn{2}{c|}{Unseen}  \\ 
\cline{2-4}
 & FF++  & Celeb-DF & WDF               \\ 
\hline
0.0     & 96.30      & 64.80  &  62.72 \\
0.1     & 97.21 & \textbf{75.24}    & 61.92             \\ 
0.2     & 96.5  & 73.35    & 64.77             \\ 
0.3     & 93.72 & 73.64    & 66.18             \\ 
0.4     & 93.63 & 71.17    & 70.18             \\ 
0.5     & 93.26 & 67.72    & 71.53             \\ 
0.6     & 91.97 & 66.19    & 72.15             \\ 
0.7     & 90.24 & 67.16    & 70.86             \\ 
0.8     & 88.63 & 68.06    & 71.32             \\ 
0.9     & 85.53 & 65.58    & \textbf{72.04 }            \\
\hline
\end{tabular}
\end{table}

Because of the different dependencies of high-frequency components in various datasets, we cannot select the specific value of $r$. Here, we propose a parameter selection method based on CAM. We calculate CAM average graph and perform the binary operation. We propose to choose parameters that could produce regions with similar forgery attention regions.
Specifically, since both the Celeb-DF dataset and the FF++ dataset use the deepfake forgery algorithm, their attention regions are similar, so we choose a smaller $r$; while the difference in the attention regions of the WDF dataset is very obvious, and we suggest to choose a larger $r$. And in the subsequent comparison experiments, we fix $r=0.1$ to ensure fairness.

\subsection{Comparison Results}

\begin{table}
\caption{Experiments trained on FF++ dataset and cross-dataset AUC(\%) evaluation on Celeb-DF dataset, WDF dataset and DFD dataset. The best results for each group are shown in bold numbers. Image/frame-based methods are marked yellow, video based methods are marked green.}\label{table:crpss}
\centering
\begin{tabular}{|c|c|c|c|c|} 
\hline
\multirow{2}{*}{Methods}        &  Source & \multicolumn{3}{c|}{Unseen}  \\ 
\cline{2-5}
& FF++  & Celeb-DF & WDF    & DFD   \\
\hline
\rowcolor{yellow!10} Xception~\cite{chollet2017Xception}         & 96.30  & 64.80     & 62.72  & 77.86 \\
\rowcolor{yellow!10} RFM~\cite{wang2021representative}              & 98.79 & 65.63    & 57.75  & -     \\
\rowcolor{yellow!10} Add-Net~\cite{zi2020wilddeepfake}          & 97.74 & 65.29    & 62.35  & -     \\
\rowcolor{yellow!10} F3-Net~\cite{qian2020thinking}           & 98.10  & 61.51    & 57.10   & 86.10  \\
\rowcolor{yellow!10} MultiAtt~\cite{zhao2021multi}         & 99.29 & 67.02    & 59.74  & -     \\
\rowcolor{yellow!10} RECCE~\cite{cao2022end}            & 99.32 & 68.71    & 64.31  & -     \\
\rowcolor{yellow!10} LTW~\cite{sun2021domain}              & 99.17 & 77.14    & 67.12  & \textbf{88.56} \\
\rowcolor{yellow!10} GFF~\cite{luo2021generalizing}              & 98.36 & 75.31    & 66.51  & 85.51 \\
\rowcolor{yellow!10} TripletNet~\cite{liang2023depth} & 99.8  & 72.30     & -      & -     \\
\rowcolor{yellow!10} TI2NET~\cite{liu2023ti2net}     & \textbf{99.95} & 68.22    & -      & 72.03 \\
\hline
\rowcolor{green!10} I3D~\cite{gu2022hierarchical}          & 95.41 & 74.11    & -     & -         \\ 
\rowcolor{green!10} TEI~\cite{gu2022hierarchical}          & 96.54 & 74.66    & -     & -         \\ 
\rowcolor{green!10}TAM~\cite{gu2022hierarchical}         & 97.04 & 67.96    & -     & -         \\ 
\rowcolor{green!10}V4D~\cite{gu2022hierarchical}        & 96.74 & 70.08    & -     & -         \\ 
\rowcolor{green!10}DSANet~\cite{gu2022hierarchical}      & 96.88 & 73.71    & -     & -         \\ 
\rowcolor{green!10}TD-3DCNN~\cite{zhang2021detecting}   & -     & 57.32    & -     & -         \\ 
\rowcolor{green!10}DoubleRNN~\cite{masi2020two}  & 93.18 & 73.41    & -     & -         \\ 
\rowcolor{green!10}ADDNet-3D~\cite{zi2020wilddeepfake} & 96.22 & 60.85    & -     & -         \\ 
\rowcolor{green!10}STIL~\cite{gu2021spatiotemporal} & 97.12 & 75.58    & -     & -         \\ 
\rowcolor{green!10}HCIL~\cite{gu2022hierarchical}        & 98.32 & 79.00       & -     & -         \\ 
\rowcolor{green!10}Forensic~\cite{li2023forensic}     & 99.88 & 61.97    & - & 83.88         \\ 

\textbf{ACMF}         & 99.30 &\textbf{ 84.02}    & \textbf{74.22} & 87.77    \\
\hline
\end{tabular}
\end{table}

To verify the effectiveness of the proposed ACMF in forgery face detection, we evaluated our proposed method on FF++, Celeb-DF, WDF, and DFD datasets. Noting
that we select videos as the unit for the following comparison experiments.

\textbf{Results on FF++ Dataset} As shown in Tabel~\ref{table:crpss}, we conducted inner-dataset experiment on the FF++ dataset, and reported AUC. We present the results of image-based methods (RFM~\cite{wang2021representative}, Add-Net~\cite{zi2020wilddeepfake}, F3-net~\cite{qian2020thinking}, MultiAtt~\cite{zhao2021multi}) and  video based methods (TD-3DCNN~\cite{zhang2021detecting}, HCIL~\cite{gu2022hierarchical}, Forensic~\cite{li2023forensic}) for comparisons. ACMF's inter-dataset result is slightly lower than existing methods, and the AUC of our method does go above 99\%. But since our method focuses on generalizability testing, we only guarantee to be competitive in inter-test experiments. However, our approach is still superior to STIL~\cite{gu2021spatiotemporal}(97.12\%), HCIL~\cite{gu2022hierarchical}(98.32\%), GFF~\cite{luo2021generalizing}(98.36\%).

\textbf{Results on Cross-dataset Evaluation} As shown in Tabel~\ref{table:crpss}, we conducted cross-dataset experiments on the Celeb-DF, WDF, and DFD datasets and reported AUC.
Many methods, while performing well within the dataset, suffer from severe performance degradation against data of unknown forgery types.
cThe AUCs of F3-net~\cite{qian2020thinking} and RFM~\cite{wang2021representative} under the WDF dataset are only 57\%. the AUCs of Forensic~\cite{li2023forensic} and TD-3DCNN~\cite{zhang2021detecting} under the Celeb-DF dataset were only 61.97\% and 57.31\%, respectively. MultiAtt~\cite{zhao2021multi} has an AUC of 67.02\% for the Celeb-DF dataset but only 59.74\% for the WDF dataset, while our method adaptively adjusts the attention region for different forged datasets and performs better in each dataset.
Our method outperforms most of the methods on the unknown forgery method as well as on the compressed WDF dataset. Specifically, ACMF can achieve about 5\% higher AUC than GFF on the Celeb-DF dataset, but the AUC of ACMF on the WDF dataset does not drop or even is much higher than other methods, while our method adaptively adjusts the attention region for different forged datasets and performs better in each dataset. Experiment results prove that the proposed ACMF could achieve satisfactory forgery detection performance even in various scenes and conditions.


\subsection{Ablation study}

The proposed ACMF framework mainly contains two designed modules: the masked frequency forgery representation (MFR) module and the forgery attention consistency (FAC) module. To reveal each module contributes to performance improvement, we conduct a comprehensive ablation study to analyze them on the FF++ and WDF datasets as shown in Table~\ref{table: Ablation}. It noted that we selected videos as the unit for the following comparison experiments.
The performance of the proposed method variants is summarized in Table~\ref {table: Ablation} on the FF++ and WDF datasets. We utilize the pure Xception network as the baseline method for a fair comparison.
Because robust features are difficult to learn, the baseline performance is poor in the inner-dataset and cross-dataset. With
the designed MFR model, the AUC of inner-dataset increased from
97.75\% to 99.2\%,  the AUC of cross-dataset increased from
65.50\% to 66.10\%. Because our designed MFR model can explore more robust forgery clues. When additionally the FAC module is utilized, the AUC of cross-dataset increased from 66.10\% to 72.60\%, achieving the best forgery detection performance. It is benefited from refining the forgery detection network focusing on similar spatial attention regions to boost performance.



\begin{table}
\caption{The ablation study to analyze the effects of different components. Specifically, we use MFR to denote the masked frequency forgery representation module and FAC to denote the forgery attention consistency module. The best results for each group are shown in bold numbers.}\label{table: Ablation}
\centering
\begin{tabular}{|c|c|c|c|c|c|} 
\hline
\multirow{2}{*}{Baseline}
& \multirow{2}{*}{MFR} & \multirow{2}{*}{FAC}  & Source   & Unseen    \\
\cline{4-5}
&    &  & FF++   & WDF    \\
\hline
\checkmark      &  -     & -   & 97.75     & 65.60  \\
\checkmark     & \checkmark      & -     & 99.20     & 72.05  \\
\checkmark   & \checkmark      & \checkmark    & \textbf{99.30}    & \textbf{74.22} \\
\hline
\end{tabular}
\end{table}




\begin{figure}
\begin{center}
\includegraphics[width=\linewidth]{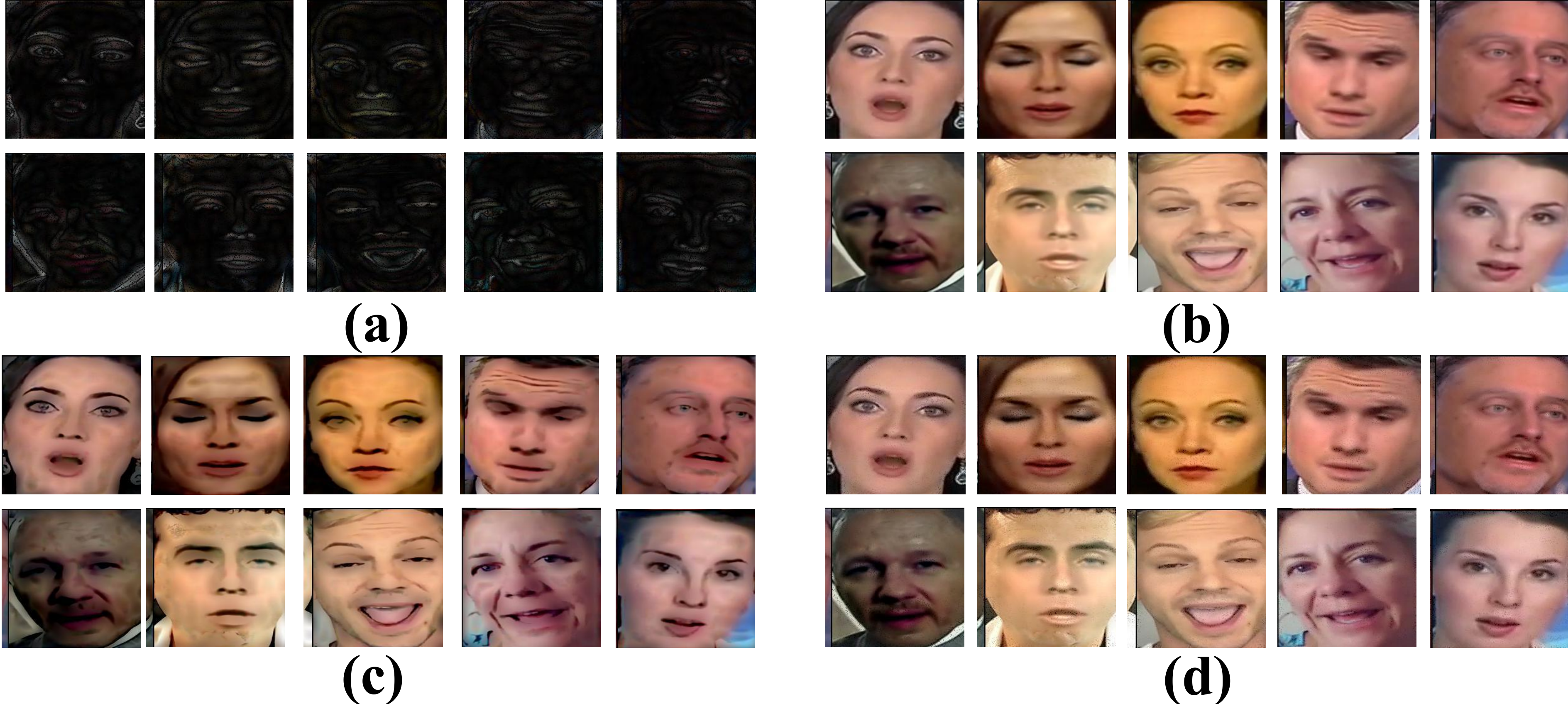} 
\caption{Visualization of Frequency Components: (a) the high-frequency component; (b) the original image; (c) the low-frequency component; (d) the processed image.}
\label{fig:train}
\end{center}  
\end{figure}

\begin{figure}
\begin{center}

\includegraphics[width=\linewidth]{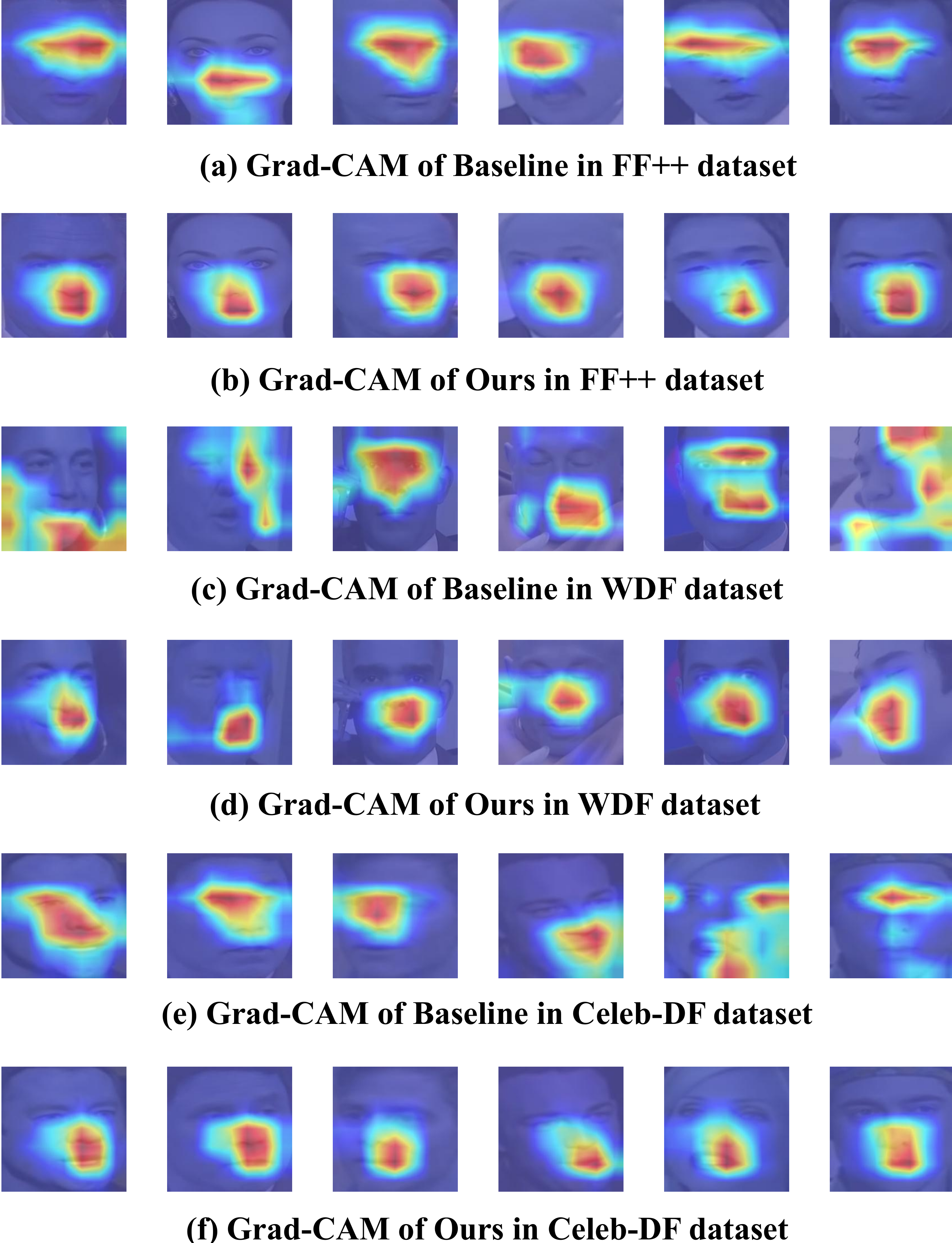} 
\caption{The Grad Class activation maps (Grad-CAM) of the Baseline and the proposed method. Images are overlayed with the important attention regions highlighted with Grad-CAM~\cite{selvaraju2017grad}. (a) is the Grad-CAM of Baseline in FF++, (b) is the Grad-CAM of the proposed method in FF++, (c) is the Grad-CAM of Baseline in Celeb-DF dataset, (d) is the Grad-CAM of the proposed method in Celeb-DF dataset,  (e) is the Grad-CAM of Baseline in WDF dataset, (f) is the Grad-CAM of the proposed method in WDF dataset. }
\label{heatmap2}
\end{center}  
\end{figure}

\subsection{Further Analysis}

\textbf{Visualization of Frequency Components.} To illustrate how our method separates low-frequency information from high-frequency information in the training image, we extracted 50\% of the frequencies and visualized the result. Figure~\ref{fig:train} compares the original image with its low-frequency and high-frequency components. The figure shows that the high-frequency information captures the texture and edge details, while the low-frequency information represents the main structure of the image, which is the area where the brightness or gray level changes gradually. 
The randomly masked image is indistinguishable from the original image to the naked eye and can help the model learn the low-frequency components that are more familiar to humans as well as the high-frequency components that are robust.

\textbf{Grad-CAM Visulization Results.} 
Many works~\cite{tariq2021one,tariq2021one,wang2021representative,miao2021learning,cao2022end} used the Grad-CAM to analyze the area of attention of the designed forgery detector, which was mainly centered on the face (around the nose). Figures~\ref{heatmap2} show Grad-CAM maps of  Baseline and Baseline+MFR. We analyze Grad-CAM maps for inner-dataset (FF++ dataset) as well as cross-dataset (WDF and Celeb-DF datasets). After masking out 50\% of the high-frequency information, our method produces Grad-CAM images that are similar in size and center around the nose, indicating robust and discriminative forgery detection cues. The Baseline network, however, generates more diverse Grad-CAM images in shape and location, implying that it may overfit specific or quality-dependent forgery features in the training data, thus compromising the generalization performance.

To analyze the changes in the region of interest of the model after adding attention consistency
regularization, we compare the Grad-CAM map with and without the addition of the forgery attention consistency model. For better visualization, we binarize the top 10\% of the forged pixels in the maps. Figures~\ref{heatmap1} show Grad-CAM maps of  Baseline+MFR and Baseline+MFR+AC. We analyze Grad-CAM maps for cross-dataset (WDF, DFD, and Celeb-DF datasets). 
We notice that the Grad-CAM attention maps with consistency constraints are more coherent across datasets than the ones without consistency constraints. 
It is because these indeed exist large domain gaps between different forgery artifact types.
Thus, it is hard to design a uniform forgery detection to accurately identify any forgery types in the real world.
The proposed FAC constraint provides a novel perspective that we can design a simple yet effective optimization strategy to process individual test videos in several epochs.
Besides, the visualization results clearly prove the proposed FAC strategy could effectively refine the detector to align them with a similar attention map, and recognition results prove this simple strategy could help boost generalization ability.

\begin{table*}
\caption{The robustness evaluation with perturbation. Specifically, we use MFR to denote the masked frequency forgery representation module and FAC to denote the forgery attention consistency module. The best results for each group are shown in bold numbers.}\label{supp: Ablation}
\centering
\begin{tabular}{|c|c|c|c|c|c|c|} 
\hline
\multirow{1}{*}{Baseline}
& \multirow{1}{*}{MFR} & \multirow{1}{*}{FAC}   & Sharpness & Brightness & Gaussian noise & Color  \\
\cline{4-5}
\hline
\checkmark      &  -     & -          & 68.81     & 64.82      & 60.58          & 70.16          \\
\checkmark     & \checkmark      & -            & 77.94     & 74.66      & 74.01          & 78.39 \\
\checkmark   & \checkmark      & \checkmark   & \textbf{82.19}    & \textbf{83.84}      & \textbf{83.42 }         & \textbf{83.40} \\

\hline
\end{tabular}

\end{table*}

\textbf{Robustness Evaluation.}
To evaluate the robustness of the proposed method in the real world, we analyze the detection performance with various perturbations similar with \cite{gu2022exploiting, yang2022deepfake}.
These perturbations include Sharpness, Brightness, Gaussian noise, and Color. 
The experimental settings for each perturbation as follows: the factor of sharpness is set as 2, the factor of brightness is set as 1.5, the mean of Gaussian noise is set as 0, the variance of Gaussian noise is set as 0.0001, factor of color is set as 2.
Table~\ref{supp: Ablation} shows the results of training on the FF++ dataset and testing on Celeb-DF~\cite{li2020celeb} dataset. 
We also apply different perturbations to simulate the real-world image degradation process. 
The Baseline algorithm suffers from a significant performance drop in the perturbed testing set. 
The designed MFR module can mitigate the overfitting problem to some extent by exploring invariant cues.
Benefited from the forgery attention consistency, the proposed algorithm could effectively focus on similar spatial attention regions to boost performance.

\begin{figure}
\begin{center}
\includegraphics[width=\linewidth]{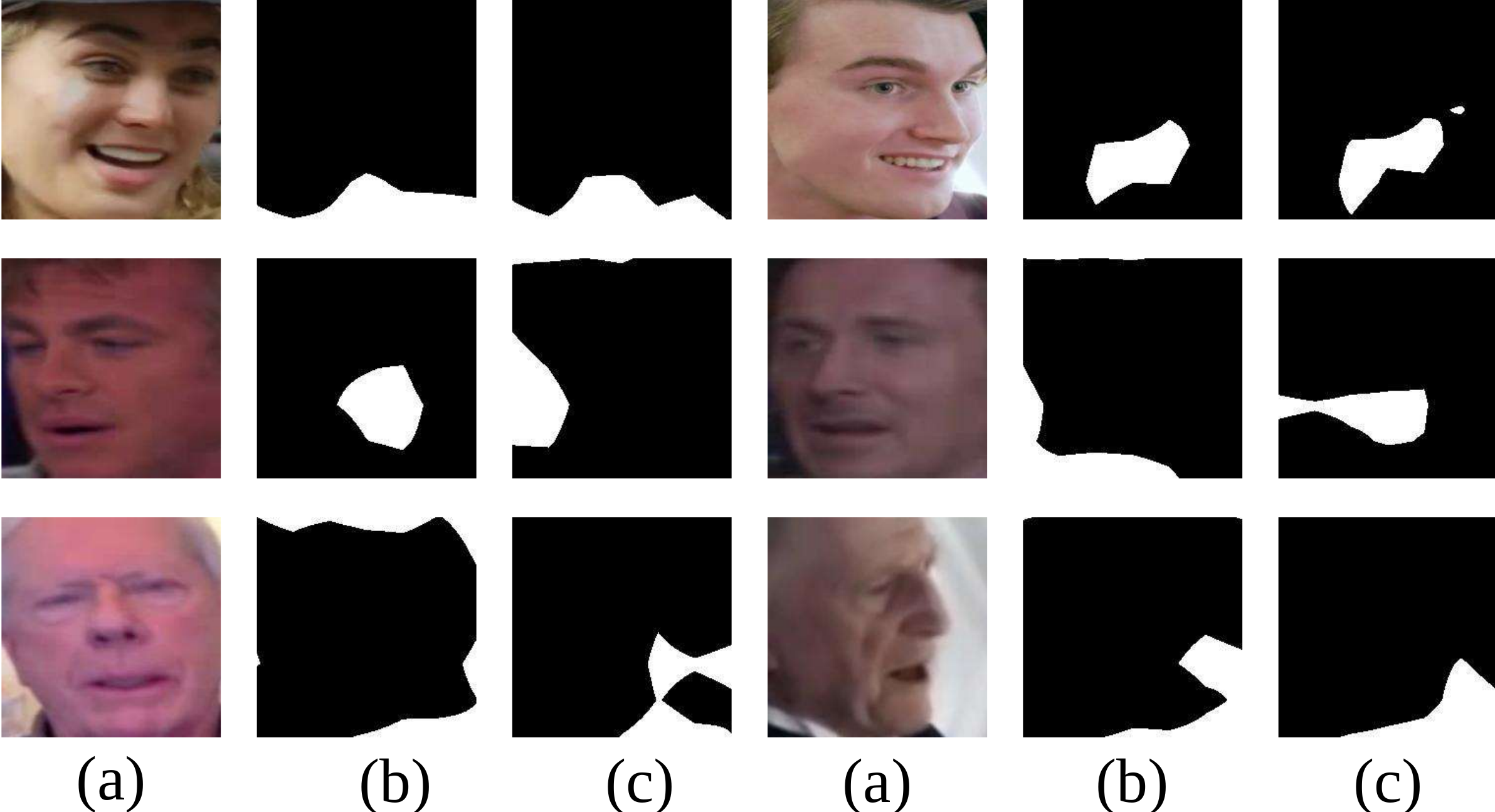} 

\caption{Failure cases visualization. (a) is original video. (b) is the Grad-CAM~\cite{selvaraju2017grad} of Baseline+MFR. (c) is the Grad-CAM of Baseline+MFR+FAC. We binarize the top 10\% of the forged pixels in the attention map.}
\label{fig:show}
\end{center}  
\end{figure}

\textbf{Failure Cases Analysis.}
Here we analyze failure cases of the proposed method.
As shown in Figure~\ref{fig:show}, some face examples are distinguished incorrectly by adding designed constraints.
The reason is that these failure samples contain various pose angles, which will degrade the forgery attention consistency and evaluation performance.
It also inspires researchers to explore stronger pose-invariant clues for generalizing forgery detection.
Meanwhile, the forgery prediction speed of the proposed algorithm is slower than other methods due to the post-processing refinement operation.

\section{Conclusion}
\label{Conclusion}

The paper proposes a novel masked frequency forgery representation and forgery attention consistency for generalizing face forgery detection. 
We find that redundant high frequency information works well in distinguishing source forgery types but performs poorly for unseen artifacts.
The masked frequency forgery representation module is proposed to explore forgery cues by erasing high frequency information randomly. 
In addition, we leverage the cross attention consistency regularization to constrain the detection network to focus on similar attention regions, which is inspired by the mentioned phenomenon.
The experimental results show that the proposed method achieves superior performance compared with the SOTA methods. 
In the future, we will explore robust face forgery clues in more complex real scenarios.

%

\bibliographystyle{IEEEtran}
\bibliography{egbib}

\end{document}